\documentclass[runningheads,a4paper]{llncs}

\usepackage{amssymb}
\usepackage{graphicx}
\usepackage{listings}
\usepackage{url}
\usepackage{arydshln}
\usepackage{float}
\urldef{\mailsa}\path|rimah.amami@yahoo.fr|
\urldef{\mailsb}\path|dorrainst@yahoo.fr|
\urldef{\mailsc}\path|n.ellouze@enit.tn|
\newcommand{\keywords}[1]{\par\addvspace\baselineskip
\noindent\keywordname\enspace\ignorespaces#1}

\begin{document}

\mainmatter  

\title{Incorporating Belief Function in SVM for Phoneme Recognition }


%
%
\author{Rimah Amami* %
\and Dorra Ben Ayed \and Nouerddine Ellouze}
%


\institute{*Institut Sup\'{e}rieur des Technologies M\'{e}dicales de Tunis (ISTMT) , Tunisia.
Department of Electrical Engineering, National School of Engineering of Tunis (ENIT), Tunisia.\\
Universit\'{e} Tunis El Manar, BP 37, Le B\'{e}lved\`{e}re 1002 Tunis, Tunisia.\\
\mailsa\\
\mailsb\\
\mailsc\\
}

%
%

\toctitle{Belief SVM for hierarchical phoneme recognition}
\tocauthor{R.Amami,D. Ayed N.Ellouze}
\maketitle

\begin{abstract}
The Support Vector Machine (SVM) method has been widely used in numerous classification tasks. The main idea of this algorithm is based on the principle of the margin maximization to find an hyperplane which separates the data into two different classes.In this paper, SVM is applied to phoneme recognition task. However, in many real-world problems,  each phoneme in the data set for recognition problems may differ in the degree
of significance due to noise, inaccuracies, or abnormal characteristics; All those problems can lead to the inaccuracies in the prediction phase.
Unfortunately, the standard formulation of SVM does not take into account all those problems and, in particular, the variation in the speech input.\\
This paper presents a new formulation of SVM (B-SVM) that attributes to each phoneme a confidence degree computed based on its geometric position in the space. Then, this degree is used in order to strengthen the class membership of the tested phoneme.
Hence, we introduce a reformulation of the standard SVM that incorporates the degree of belief.
Experimental performance on TIMIT database shows the effectiveness of the proposed method B-SVM on a phoneme recognition problem.
\end{abstract}

\section{Introduction}

\label{intro}

Support Vector Machine (SVM) was, at first, introduced by Vladimir  \cite{Vapnik} for a binary classification tasks in order to construct, in the input space, the decision functions  based on the theory of Structural Risk Minimization, (\cite{Cortesal} and \cite{Scholkopfal}).
Afterwards, SVM has been extended to support either the multi-class classification and  regression tasks.
SVM consists of constructing one or several hyperplanes in order to separate the data into the different classes. Nevertheless, an optimal hyperplane must be found in order to separate accurately  the data into two classes.\\
\cite{Cortesal} defined the optimal hyperplane as the decision function with maximal margin. Indeed, the margin can be defined as the shortest distance from the separating hyperplane and the closest vectors to the couple of classes.
The application of SVM to the automatic speech recognition (ASR) problem has shown a competitive performance and accurate recognition rates.
In the sound system of a language, a phoneme is considered as the smallest distinctive unit which is able to communicate a possible meaning. Thus, the success of the phoneme recognition task is important to the development of language systems.
Nevertheless, during the signal acquisition process, the speech signal may be affected by the speaker characteristics such as his gender, accent, and style of speech.  Also, there are other external factors which can admittedly have an impact on the speech recognition such as the noise coming from a microphone or the variation in the vocal tract shape.\\

The standard formulation of SVM may not  determine accurately the identity of the tested phoneme.  Indeed, the speech signal is accompanied by all sorts of unpleasant variations during the acquisition.
Those variations affect badly the recognition rates since the recognition mechanism may not be taken into account those changes in the phoneme data.
For example, in the real-application problems, the English pronunciation differences and the differences in accents may lead to increase significantly the error rate of any learning algorithm since all phoneme data are handled identically.
Thus, the standard SVM may find an optimal hyperplane without considering the influences of the differences accompanied by the speech signals. Thus, the identified optimal hyperplane can lead to loss of accuracies.\\
In this paper, we propose a novel approach in order to incorporate a belief function into the standard SVM algorithm which involves integrating confidence degree of each phoneme data.
To fulfill this new formulation, we have, beforehand, compute the geometric distance between the centers of each possible class of the tested phoneme. Indeed, the benefit of hybrid approaches relies in their support to the decision-making  and their ability to confirm the robustness of the recognition system \cite{1}, \cite{2}.
The experimental results with all phoneme datasets issued from the TIMIT database \cite{Garofoloal} show that the B-SVM outperforms the standard SVM and produces a better recognition rates.
The rest of this paper is organized as follows: Section \ref{1} presents an overview of the method Support Vector Machines (SVM). Section \ref{2} presents the steps of the phoneme processing and the problems which accompanying the speech processing.
Section \ref{bsvm} presents the new formulation B-SVM algorithm; Section \ref{sys} describes the hierarchical phoneme recognition system; Section \ref{res} presents the experimental results and a comparison between the standard SVM and B-SVM in a multi-class phoneme recognition problem. The final section is the conclusion.

\section{Support Vector Machines}\label{1}
The Support Vector Machines (SVM) is a learning algorithm for pattern recognition and regression problems \cite{Singer} whose approaches the classification problem as an approximate implementation of the Structural Risk Minimization(SRM) induction principle \cite{Cortesal}.\\
 SVM approximates the solution to the minimization problem of SRM through a Quadratic Programming optimization.
It aims to maximize the margin  which is the distance from a separating hyperplane to the closest positive or negative sample between classes.\\
Hence the hyperplane that optimally separates the data is the one that minimises:
\begin{equation}\label{}
  \frac{1}{2}\|w^{ij}\|^2+C\sum_{i=1}^m\xi^{ij}
\end{equation}
Where $C$ is a penalty to errors and $\xi$ is a positive slack variable which measures the degree of misclassification.\\
subject to the constraints:
\begin{eqnarray}
  (w^{ij})  \phi(x_t)+b^{ij} \geq 1-\xi^{ij},  \mbox{si}  ~~    y=i \nonumber \\
    (w^{ij})  \phi(x_t)+b^{ij} \leq 1-\xi^{ij},  \mbox{si}  ~~  y=j \nonumber \\
  \xi^{ij}_t\geq 0
\end{eqnarray}

For the phoneme classification, the decision function of SVM is expressed as:
\begin{equation}
  f(x)=sign(\sum_{i=1}^m    \alpha_i   y_i   K(x_i,x) +b)
\end{equation}
The above decision function gives a signed distance from a phoneme x to the hyperplane.

However, when the data set is linearly non-separable, solving the parameters of this decision function becomes a quadratic programming problem. The solution to this optimization problem can be cast to the Lagrange functional and the use of Lagrange multipliers $\alpha_i$, we obtain the Lagrangian of the dual objective function:
\begin{equation}
L_d = \max_{\alpha_i} \sum_{i=1}^m \alpha_i - \sum_{i=1}^m \sum_{j=1}^m  \alpha_i  \alpha_j y_i y_j K(x_i, x_j).
\end{equation}
where $K( x_i, x_j)$ is the kernel of data $x_i$ and $x_j$ and the coefficients $\alpha_i$ are the lagrange multipliers and are computed for each phoneme of the data set. They must be maximised with respect to $\alpha_i\geq 0$. It must be pointed out that the data with nonzero coefficients $\alpha_i$ are called support vectors. They determine the decision boundary hyperplane of the classifier.\\
Moreover, applying a  kernel trick that maps an  input vector into a higher dimensional feature sapce, allows to SVM to approximate a non-linear function \cite{Cortesal} and \cite{Li08}.
In this paper, we use SVM with the radial basis function kernel (RBF).This kernel choice was made after doing a case study in order to find the suitable kernel with which SVM may achieve good generalization performance as well as the parameters to use \cite{amamijdcta}.
Based on this principle, the SVM adopts a systematic approach to find a linear function that belongs to a set of functions with lowest VC dimension (the Vapnik–Chervonenkis dimension measure the capacity of a statistical classification algorithm).
\section{Phoneme processing}\label{2}

Speech recognition is the process of converting an acoustic signal, captured by a microphone , to a set of words, syllables or phonemes. The speech recognition systems can be used for applications such as mobiles applications, commands, control, data entry, and document preparation.
The steps of the speech processing are described in the figure 1:\\
\begin{figure}
\centering
\includegraphics[scale=0.65]{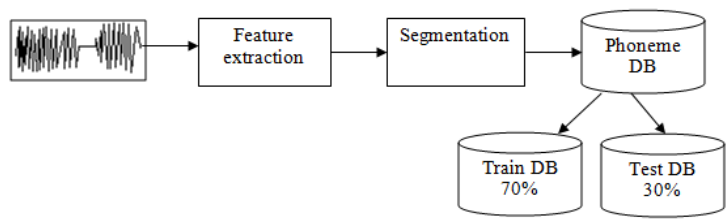}
\caption{Phoneme processing steps.}
\label{fig:examplee}
\end{figure}

The phoneme processing consists, first, on converting the speech captured by a microphone to a  sequence of feature vectors. Then, a segmentation step is applied consisting on converting the continued speech signal to a set of units such as phonemes. Once the train and test data sets are prepared, a classifier is applied to classify the unknown phonemes.
However, the phoneme recognition systems can be characterised by many parameters and problems which have the effect of making the task of recognition more difficult. Those factors can not be taking into account by the classifier since their accompanying the captured speech.\\
In fact, the speech contains disfluencies, or periods of silence, and is much more difficult for the classifier to recognise than speech periods.
In the other hand, the speaker is not able to say phrases in the same or similar manner each time. Thus, the phoneme recognition systems learn barely to recognize correctly  the phoneme.  The speaker's voice quality, such as volume and pitch, and breath control should also be taken into account since they distorted the speech. Hence, the physiological elements must be taken into account in order to construct a robust phoneme recognition.\\

Regrettably, the classifier is not able to take into account all those external factors which are inherent in the signal speech  in the recognition process which may lead to a confusion inter-phonemes problem. In this paper, we propose to incorporate a confidence degree which will help the standard classifier SVM to find the optimal hyperplane and classify the phoneme into its class.

\section{Belief SVM (B-SVM)}\label{bsvm}

The formulation of the proposed method B-SVM is described in three steps; the first step consists of computing the Euclidean distance $d(Y_i,X_i)$ between the center of the different classes and the phoneme to be classified $x_i$. The second step is to compute the confidence degree of the membership of the phoneme $x_i$ into the class $y_i$. Then, those confidence degrees are incorporated into SVM to help to find the optimal hyperplane.
\subsection{Geometric distance}
We propose to calculate the geometric distance between $X_i$ and the center of the class $CY_i$ where $i\in(1,\ldots,k)$.   We consider that there is a possibility to which the phoneme $X_i$ belongs to one of the classes $Y_i$.
The geometric distance noted $d(CY_i,X_i)$ is calculated using euclidian distance. \\
The higher value of $d(CY_i,X_i)$ is assigned to the most distant class $Y_i$ from the phoneme $X_i$ and the lower  value is associated with the closer class to the phoneme $X_i$.

\subsection{Confidence degree}

This step consists on calculating the confidence degree $m_i(X)$ of each phoneme $X_i$. It tells the possibility that $X_i$ belongs to the class $Y_i$.
This proposed algorithm  allows the generation of confidence degree for each phoneme:\\
\medskip
\noindent
{\it Calculate confidence degrees $m_i(X_i)$}
\begin{lstlisting}[mathescape]
   begin
     Set of phoneme samples with lables $\{(X_1,Y_1),\ldots, (X_n,Y_k)\}$;
   Initialize confidence degree $m_i$ of samples:
   1 if $X_i$ in the ith class, 0 Otherwise;
    $C_i $:= Center of the ith class;
    $m_i(X) := 1/d(C_i,X_i)$
    end.
\end{lstlisting}
\noindent

\subsection{Formulation of belief SVM}
In a space where the data sets are not linearly separable and a multi-class classification problem, SVM constructs $k(k-1)/2$ classifiers for the training data set. In order to convert the multi-class problem into multiple binary  problems, the approach one-against-one is used.

In the proposed B-SVM, we incorporate the confidence degree of each phoneme samples into the constraints since the identity is not affected by a scalar multiplication. We normalized the hyperplane to satisfy:

 \begin{eqnarray}
  \textbf{m(x)}(w^{ij})^T  \phi(x_t)+b^{ij} \geq 1-\xi^{ij}_t,  \mbox{if}  ~~    y_t=i \nonumber \\
    \textbf{m(x)}(w^{ij})^T  \phi(x_t)+b^{ij} \leq 1-\xi^{ij}_t,  \mbox{if}  ~~  y_t=j \nonumber \\
  \xi^{ij}_t\geq 0
\end{eqnarray}
In fact, the incorporation of the confidence degree allows to to reduce the restrictions when the phoneme have a high degree into the class.

In the other hand, the dual representation of the standard SVM allows to maximise the $\alpha_L$ of each phoneme. Thus, with high value of the confidence degree, the subject to $\alpha_i\geq 0$ can be easily satisfied which allows to consider this one as support vector which be helping to decide on the hyperplane.\\
In the proposed B-SVM, we optimize this formulation to obtain a new dual representation:
\begin{equation}
L_d = \max_{\alpha_i} \sum_{i=1}^m \alpha_i - \sum_{i=1}^m \sum_{j=1}^m \textbf{$m(x_i)$} \textbf{$m(x_j)$}\alpha_i  \alpha_j y_i y_j \Phi(x_i)  \Phi(x_j).
\end{equation}

In the standard SVM, the class $Y_i$ of a phoneme $X$ is determined by the sign of the decision function.
In the proposed B-SVM, the new decision function thus becomes:
\begin{equation}
  \sum_{i=1}^m   \textbf{m(x)} \alpha_i   y_i   \Phi(x_i) +b
\end{equation}

This new formulation will help for the decision making on the sign of phoneme in order to  classify  into its class.
\section{Hierarchical phoneme recognition system}\label{sys}

The architecture of our Hierarchical phoneme recognition systems is described in the figure 2:
\begin{figure}
\centering
\includegraphics[scale=0.65]{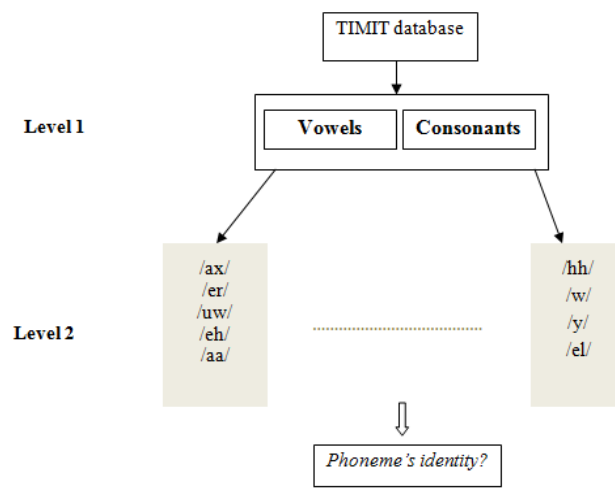}
\caption{Hierarchical phoneme recognition system.}
\label{fig:example}
\end{figure}

The recognition system proceeds as follows: (1) conversion from the speech waveform to a spectrogram (2) transforming the spectogram to a Mel-frequency cepstral coefficients (MFCC) spectrum using the Spectral  analysis (3) segmentation of the phoneme data sets to sub-phoneme data sets (4) initiating the phoneme recognition at the first level  of the system using B-SVM to recognize the class of the unknown phoneme (vowels or consonant) (5) and, finally, initiate the phoneme recognition at the second level of the system using B-SVM to recognize the identity of the unknown phoneme (i.e. aa, ae, ih , etc) \cite{AMAMI}.

 For the proposed recognition system, we have used the  MEL frequency cepstral coefficients (MFCC) feature extractor in order to convert the speech waveform to a set of parametric representation.\\
Davis and Mermelstein  were the first who introduced the MFCC concept for automatic speech recognition \cite{DavisMermelstein}. The main idea of this algorithm consider that the MFCC are the cepstral coefficients calculated from the mel-frequency warped Fourier transform representation of the log magnitude spectrum.
The Delta and the Delta-Delta cepstral coefficients are an estimation of the time derivative of the MFCCs. Including the temporal cepstral derivative aim to improve the performance of speech recognition system.\\
Those coefficients have shown a determinant capability to capture the transitional characteristics of the speech signal that can contribute to ameliorate the recognition task.
The experiments using SVM are done using LibSVM toolbox \cite{ccChang}. The table 1 recapitulate our main choice of experiments conditions:
\begin{table}[H]
  \centering
  \caption{Experimental setup}
\begin{tabular}{l|l}
\hline
   \textbf{Method} & SVM\\
 \textbf{ $\gamma$} & 1/117 \\
  \textbf{Cost} & 10 \\
  \textbf{Kernel trick} & RBF \\
  \textbf{Windowing} & 3-middle aligned Windows \\
  \textbf{Corpus} & TIMIT \\
  \textbf{Dialect} & New England \\
  \textbf{Frame rate }& 125/s \\
   \textbf{Features technique} & MFCC \\
  \textbf{Features number} & 39 \\
  \textbf{Sampling frequency} & 16ms\\
  \hline
\end{tabular}
  \label{tab:tablee1}%
\end{table}%

It should be noted that, for the  nonlinear B-SVM method, we chose the RBF  Kernel  and the one-against-one  strategy to carry out a multi-class SVM classification.
Furthermore, the input speech signal is segmented into frames of 16 ms with optional overlap of $1/3 \sim 1/2$ of the frame size.
If the sample rate is 16 kHz and the frame size is 256 sample points, then the frame duration is $16 ms$. In addition, the frame rate is $125$ frames per second. Each frame has to be multiplied with a Hamming window in order to keep the continuity of the first and the last points in the frame.

\section{Experimental results}\label{res}
The table 2 shows  prediction accuracies at both first and second levels of the hierarchical recognition system using seven different phoneme classes.
\begin{table}[H]
  \centering
  \caption{Accuracies of B-SVM and standard SVM}
    \begin{tabular}{l|l|l|l||l|l|l}
    \hline
 \textbf{Method}& \multicolumn{3}{c||}{\textbf{B-SVM}}& \multicolumn{3}{|c}{\textbf{Standard SVM}} \\

    & \textbf{Acc.}& \textbf{Precision}& \textbf{Recall}&\textbf{Acc.}& \textbf{Precision}&\textbf{Recall} \\
    &\textbf{\%}&\textbf{\%}&\textbf{\%}&\textbf{\%}&\textbf{\%}&\textbf{\%}\\
    \hline
    \hline
   \textbf{Level 1:} &95&97&92&  93&89&88\\
    \hline
    \textbf{Level 2:}  &84&83&80 &78 &75&73\\
\hdashline
    \textbf{Vowels} & 83&86& 82&76&77&71\\
    \textbf{Occlusives} & 85&88&82 &82&86&81\\

    \textbf{Nasals} & 80&78&69 &75&63&60\\
    \textbf{Fricatives} & 87&76&78& 83&69&70\\

    \textbf{Semi-vowels} & 87&91&91& 84&91&87\\

    \textbf{Silences} & 83&71&69&75&62&68 \\

    \textbf{Affricates} & 83& 88&88&71&78&77\\
    \hline

    \end{tabular}%
  \label{tab:ressys3}%
\end{table}%
To investigate the accuracy of the proposed method B-SVM, we applied the standard SVM and B-SVM to Timit database. It must be pointed out that for the prediction, we used a test samples which were not included in the training stage.
We compare the performance of both methods and we note that the performance of B-SVM is better  than the standard SVM for all data sets used.

Thus, the following results in the table 1 provides a summary through which we note that  the proposed B-SVM shows a remarkable improvement over standard SVM.
\section{Conclusion}

In our paper, we have proposed a new formulation of SVM using the confidence degree for each object. We have, also, built an hierarchical phoneme recognition system.\\
The new method B-SVM seems to be more effective than the standard SVM for all tested data sets. The new formulation of SVM succeeded in improving phoneme recognition since the allocation of belief weights for each phoneme have the ability for modeling the similarity between phonemes in order to reduce the confusions inter-phonemes.
We compare the performance of both methods and we note that the performance of B-SVM is better  than the standard SVM for all data sets used.

\end{document}